\documentclass[10pt,twocolumn,letterpaper]{article}

\usepackage[pagenumbers]{cvpr} 

\definecolor{cvprblue}{rgb}{0.21,0.49,0.74}
\usepackage[pagebackref,breaklinks,colorlinks,allcolors=cvprblue]{hyperref}

\def\paperID{11441} 
\def\confName{CVPR}
\def\confYear{2026}

\title{From 2D Alignment to 3D Plausibility: Unifying Heterogeneous 2D Priors and Penetration-Free Diffusion for Occlusion-Robust Two-Hand Reconstruction}

\author{Gaoge Han\textsuperscript{1,2} \quad Yongkang Cheng\textsuperscript{1,2}\quad Zhe Chen\textsuperscript{4} \quad Shaoli Huang\textsuperscript{1,\textbf{*}}\quad Tongliang Liu\textsuperscript{2,3}\\
\textsuperscript{1}AgiBot \\
\textsuperscript{2}Mohamed bin Zayed University of Artificial Intelligence \\
\textsuperscript{3}The University of Sydney \\
\textsuperscript{4} La Trobe University 
}

\begin{document}
\maketitle

\footnotetext{ \textbf{*} Corresponding author: Shaoli Huang. \textsuperscript{4} Dr. Zhe Chen is also affiliated with Cisco - La Trobe Centre for Artificial Intelligence and Internet of Things}
\begin{abstract}
Two-hand reconstruction from monocular images is hampered by complex poses and severe occlusions, which often cause interaction misalignment and two–hand penetration. We address this by decoupling the problem into 2D structural alignment and 3D spatial interaction alignment, each handled by a tailored component.
For 2D alignment, we pioneer the attempt to unify heterogeneous structural priors (keypoints, segmentation, and depth) from vision foundation models as complementary structured guidance for two-hand recovery. Instead of extracting priors prediction as explicit inputs, we propose a fusion-alignment encoder that absorbs their structural knowledge implicitly, achieving foundation-level guidance without foundation-level cost. For 3D spatial alignment, we propose a two-hand penetration-free diffusion model that learns a generative mapping from interpenetrated poses to realistic, collision-free configurations. Guided by collision gradients during denoising, the model converges toward the manifold of valid two-hand interactions, preserving geometric and kinematic coherence. This generative formulation approach enables physically credible reconstructions even under occlusion or ambiguous visual input. Extensive experiments on InterHand2.6M and HIC show state-of-the-art or leading performance in interaction alignment and penetration suppression. Project: https://gaogehan.github.io/A2P/
\end{abstract}

\section{Introduction}
\label{sec:intro}

3D two-hand recovery aims to reconstruct both hands of a person in 3D space, a crucial capability for emerging applications in 3D character animation, AR/VR, and robotics. Large-scale hand datasets~\citep{moon2020interhand2,moon2024dataset} have accelerated progress across lines of work that scale data~\citep{pavlakos2024reconstructing}, strengthen backbones~\citep{lin2021mesh,pavlakos2024reconstructing}, and model inter-hand relations with attention~\citep{li2022interacting,yu2023acr,lin20244dhands}. Parallel advances show the promise of foundation-model priors and generative priors for 3D recovery: in human reconstruction, WHAM~\cite{shin2024wham} leverages 2D keypoint models as motion priors, TRAM~\cite{wang2024tram} exploits segmentation and depth, and BUDDI~\citep{muller2024generativeprior} uses diffusion as a generative prior. These trends indicate that structured 2D cues inferred by vision foundation models (e.g., keypoints, segmentation masks, and depth) can provide valuable guidance for 3D hand reconstruction.

Directly applying such priors to two-hand reconstruction, however, is non-trivial. Fine-tuning large 2D encoders and handling multiple task prediction is computationally heavy and leaves 2D–3D feature alignment ambiguous; under mutual hand occlusion, 2D cues can be unreliable; and while 3D generative priors (e.g., diffusion) can model interactions, they require accurate alignment to observations and otherwise drift to implausible states. Moreover, existing two-hand methods~\cite{moon2023bringinginterwild,yu2023acr,lin20244dhands} typically lack dedicated mechanisms for alignment, leading to spatial inconsistencies, unnatural interactions, and interpenetration artifacts. We therefore decouple the problem into two complementary alignment stages: 2D structural alignment and 3D spatial interaction alignment, each addressed by a tailored component.

\begin{figure*}[t]
\begin{center}
\includegraphics[width=.9\textwidth]{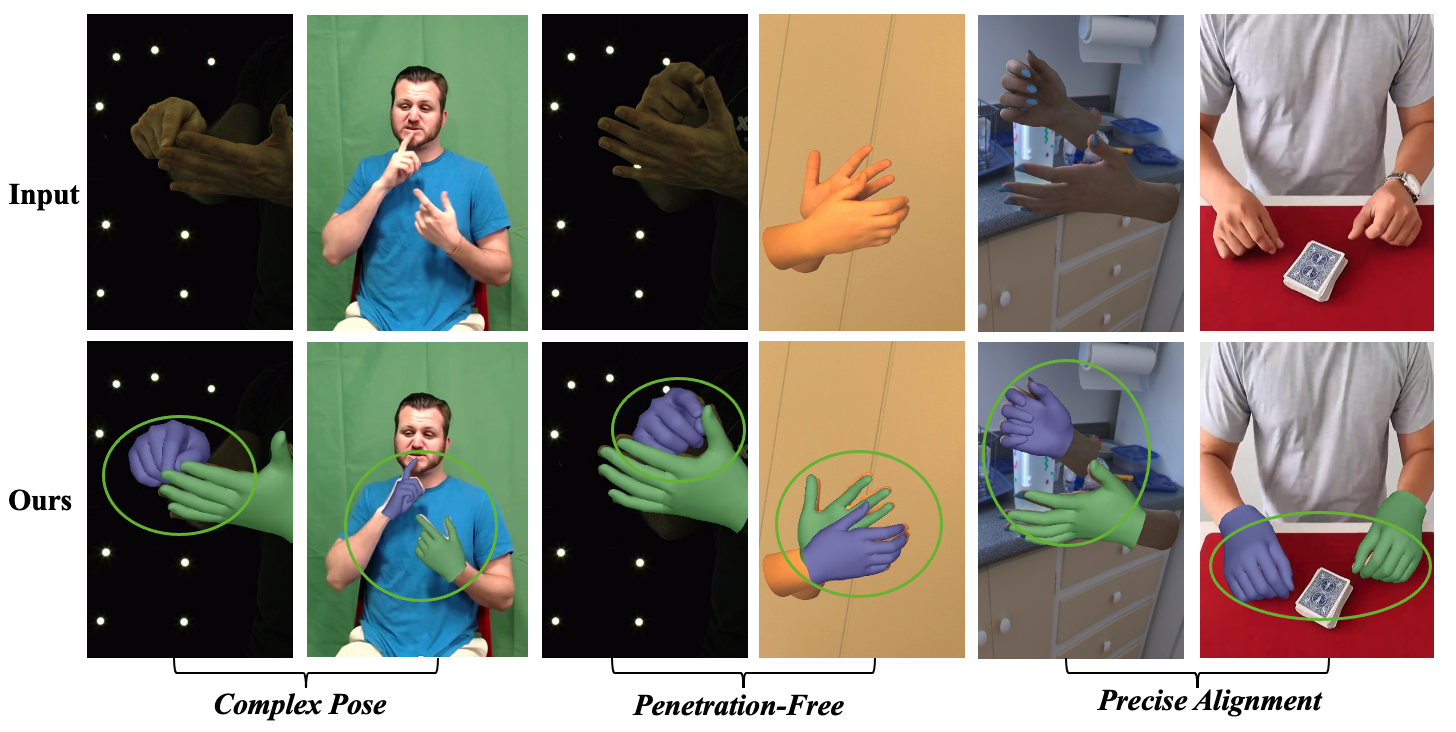}
\end{center}
\caption{Two-hand recovery on InterHand2.6M (1st, 3rd columns), Re:InterHand (4th, 5th columns), and In-the-Wild (2nd, 6th columns).}
\label{fig:teaser}
\end{figure*}

To this end, we decouple two-hand recovery into 2D alignment and 3D spatial alignment and couple them in a unified pipeline. This progressive design directly targets the root causes of failure ambiguous 2D–3D correspondence and penetration, yielding occlusion-resistant two-hand reconstruction.

In the 2D stage, we are, to our knowledge, the first to unify multiple 2D structural priors, including keypoints, segmentation, and depth, from vision foundation models for two-hand recovery.
We introduce a lightweight Fusion Alignment Encoder (FAE) that integrates these heterogeneous cues in a compact, learnable way. Instead of relying on explicit prior predictions, the FAE implicitly learns fused prior features from vision foundation model's~\cite{khirodkar2024sapiens} latent outputs during training to capture consistent geometric and semantic structure. This process removes the need to run foundation models to predict multiple prior tasks, while enabling the network to internalize their complementary reasoning into a unified representation.
At inference, all foundation encoders are removed, allowing encoder-free deployment that maintains multi-prior accuracy with greatly improved efficiency.

In the 3D stage, we propose a two-hand penetration-free diffusion model that learns a generative mapping from interpenetrated poses to physically plausible, penetration-free configurations. While diffusion-based methods such as InterHandGen~\cite{lee2024interhandgen} mainly serve as output regularizers without explicitly modeling 3D spatial interactions, they often fail to fully resolve inter-hand penetrations.
Likewise, CNN-based interaction frameworks~\cite{zuo2023reconstructing} rely heavily on image-level features and lack strong geometric grounding, resulting in limited 3D consistency and unstable contact reconstruction.
To overcome these limitations, our diffusion-based spatial alignment module integrates multi-prior 2D evidence with explicit, conditioned de-penetration in 3D, enabling direct learning of feasible interaction manifolds. Furthermore, to enhance de-penetration capability, we incorporate collision gradient guidance during denoising.

Overall, this two-stage design aligns informative 2D priors and enforces 3D interaction plausibility, yielding geometrically accurate and interaction-consistent two-hand reconstructions even in the presence of occlusions, as shown in Fig.~\ref{fig:teaser}. We demonstrate robust performance across diverse scenes and interaction poses.

Our key contributions can be summarized as follows.

\begin{itemize}
\item We make the first attempt to unify heterogeneous structural priors including keypoints, segmentation, and depth for two-hand recovery through a lightweight fusion and alignment encoder used only during training, which removes heavy encoders at inference while maintaining high accuracy.

\item We introduce the first two-hand penetration-free diffusion model that learns a generative mapping to produce physically plausible and penetration-free reconstructions, achieving robust recovery even under occlusion.

\item These stages jointly address 2D and 3D alignment, enabling occlusion-aware and realistic two-hand reconstruction with state-of-the-art results on InterHand2.6M, HIC, and FreiHAND, supported by ablations demonstrating the effectiveness of multi-prior 2D alignment and diffusion-based interaction modeling.
\end{itemize}

\section{Related Work}
\label{sec:rel}

\subsection{3D Hand Recovery}
With the introduction of some high-quality hand datasets, recovering 3D hand MANO~\citep{romero2017embodied} parameters from monocular input images has recently achieved remarkable advancements.

{\bf Single-Hand Recovery:} METRO~\citep{zhang2019end} employs a convolutional neural network to extract a single global image feature and performs position encoding by repeatedly concatenating this image feature with the 3D coordinates of a mesh template. MeshGraphormer~\citep{lin2021mesh} introduces a graph-convolution enhanced transformer to effectively model both local and global interactions. AMVUR~\citep{jiang2023probabilistic} proposes a probabilistic approach to estimate the prior probability distribution of hand joints and vertices.  Zhou et al.~\citep{zhou2024simple} simplifies the process by decomposing the mesh decoder into a token generator and a mesh regressor, achieving high performance and real-time efficiency through a straightforward yet effective baseline. HaMeR~\citep{pavlakos2024reconstructing} highlights the significant impact of scaling up to large-scale training data and utilizing high-capacity deep architectures for improving the accuracy and effectiveness of hand mesh recovery. {\bf Two-Hand Recovery:} IntagHand~\citep{li2022interacting} propose a GCN-based network to reconstruct two interacting hands from a single RGB image, featuring pyramid image feature attention (PIFA) and cross hand attention (CHA) modules to address occlusion and interaction challenges. InterWild~\citep{moon2023bringinginterwild} bridges MoCap and ITW samples for robust 3D interacting hands recovery in the wild by leveraging single-hand ITW data for 2D scale space alignment and using geometric features for appearance-invariant space. ACR~\citep{yu2023acr} explicitly mitigates interdependencies between hands and between parts by leveraging center and part-based attention for feature extraction. 4DHands~\citep{lin20244dhands} handles both single-hand and two-hand inputs while leveraging relative hand positions using a transformer-based architecture with Relation-aware Two-Hand Tokenization (RAT) and a Spatio-temporal Interaction Reasoning (SIR) module. 

Although these methods have generally achieved competitive results in hand pose and shape reconstruction, their performance in finer details is still lacking.

\subsection{Integrating Task-Related Prior}
Recent advances have demonstrated that incorporating task-related prior knowledge can significantly enhance performance in various visual tasks. 

{\bf 2D Priors}~\citep{xiu2023econ,zhang2023adding,shin2024wham}: ECON~\citep{xiu2023econ} takes as input an RGB image and is conditioned on the rendered front and back body normal images in human digitization task. This strategy allows it to excel at inferring high-fidelity 3D humans in loose clothing and challenging poses. For text-to-image generation, ControlNet~\citep{zhang2023adding} has also successfully utilized different types of conditional inputs, such as sketches, depth maps, and segmentation maps. It has successfully achieved the generation of images aligned with these conditional guides using a pretrained text-to-image diffusion model. For the 3D human motion estimation task, WHAM~\citep{shin2024wham} uses human 2D key points to extract motion features as inputs for both the Motion Decoder and Trajectory Decoder. This approach achieves more robust and stable 3D human motion estimates in global coordinates. {\bf Generative Prior}~\citep{zuo2023reconstructing,lee2024interhandgen,han2024hutumotion,han2025reindiffuse,cheng2025conditional}: Zuo et al.~\cite{zuo2023reconstructing} captured interaction priors in the latent space of a VAE and applied them to interacting hand reconstruction, effectively estimating plausible hand poses. InterHandGen~\cite{lee2024interhandgen} trained a cascaded two-hand generation model, which serves as a generative prior to formulate a loss regularizer for addressing the challenge of two-hand reconstruction. 

In this paper, we attempt to acquire multimodal 2D hand priors as 2D domain constraints from foundation models and utilize a two-hand diffusion-based interaction prior to respectively address 2D and 3D alignment challenges in two-hand parameter estimation.

\begin{figure*}[h]
\begin{center}
\includegraphics[width=1.\textwidth]{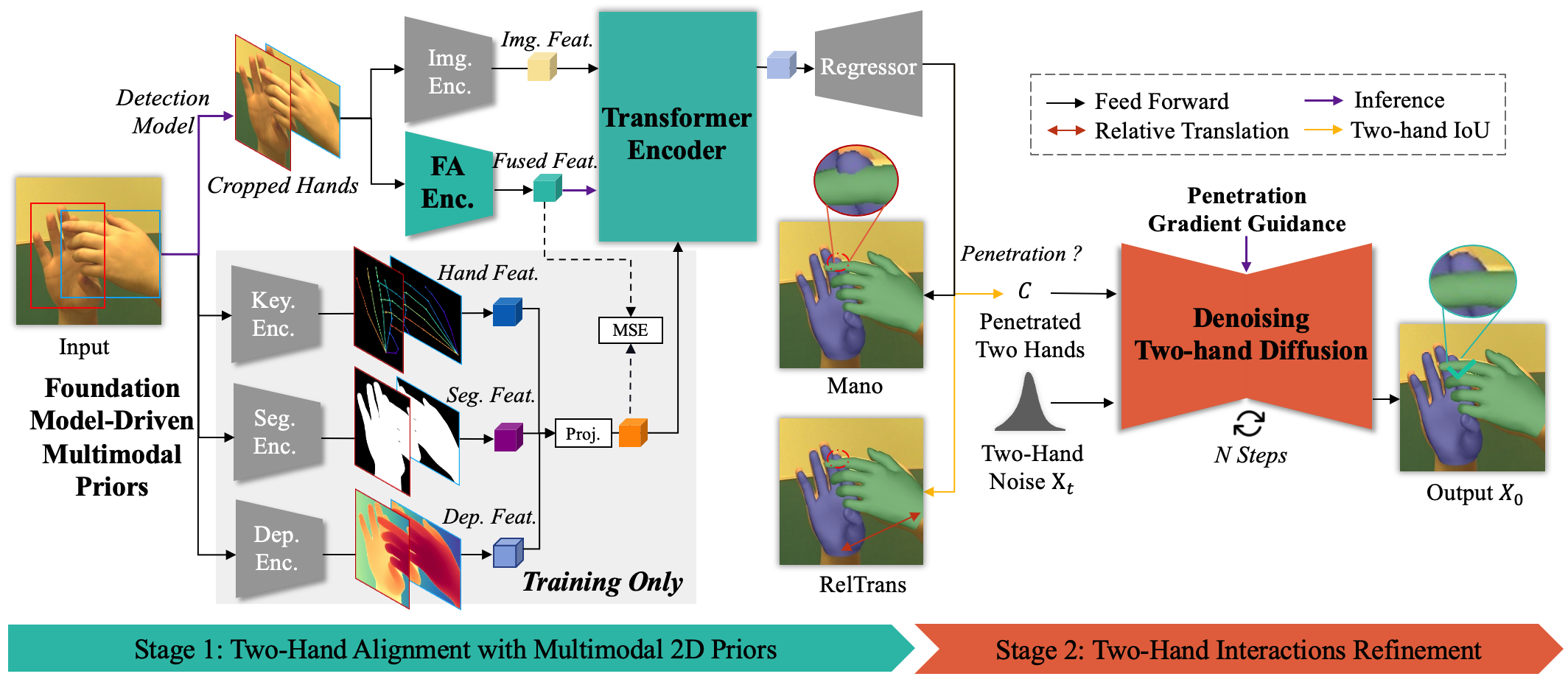}
\end{center}
 \caption{ The overall pipeline of our proposed method. ``Feat.'', ``Proj.'', ``Enc.'', ``FA'', ``Key.'', ``Seg.'', ``Pen.'' and ``RelTrans" are abbreviations for ``Feature'', ``Projection'', ``Encoder'', ``Fusion Alignment'', ``key points'', ``Segmentation'', ``Penetration'' and ``Relative Translation", respectively. $c$ denotes the condition of penetrated two hands. The purple arrow path will be activated during inference, while the yellow arrow path will be activated when the Intersection over Union (IoU) of both hands is greater than 0.}

\label{fig:pipline}
\end{figure*}

\section{Method}

This section elaborates the two-stage technical framework of our two-hand reconstruction method. As depicted in Fig. \ref{fig:pipline}, our method introduces two key innovations beyond conventional two-hand estimation pipelines: 1) Two-Hand Alignment with Multimodal 2D Priors: integration of multimodal 2D priors during training for two-hand alignment by the fusion alignment encoder, followed by 2) Two-Hand Interactions Refinement: an interaction-aware refinement process using our proposed two-hand diffusion model.

\subsection{Two-Hand Multimodal 2D Priors Alignment}
Existing approaches typically process monocular hand images by extracting visual features through a backbone network to directly regress MANO~\cite{romero2017embodied} parameters. In contrast to this standard pipeline that relies solely on image features, our method additionally incorporates structured guidance from local key points to depth cues to establish more accurate  two-hand pose and shape alignment. To obtain robust multimodal priors, we employ the human-centric vision foundation model Sapiens~\cite{khirodkar2024sapiens}, which handles 2D key points, segmentation, and depth tasks.

{\noindent \bf 2D Hand Key Points Prior.} Inspired by WHAM~\cite{shin2024wham}, which extracts whole-body 2D key points to identify human features, we focus on extracting 2D hand key points as hand features, distinct from joint-level features. These 2D key points provide precise locations of critical hand features, such as joints and fingertips, enabling a more accurate understanding of hand poses. For extracting 2D key point features, unlike WHAM, which employs an MLP and requires additional 2D-to-3D pretraining, our method eliminates these extra pre-training steps and aligns them in image space, creating a more efficient framework for integrating diverse types of information.

{\noindent \bf Two-Hand Segmentations Prior.} 
Segmentation maps offer pixel-level details for precise hand localization and background removal, reducing noise. They enable models to extract hand features more effectively by focusing on segmented regions. Notably, while heavy hand interleaving may make 2D keypoint predictions unreliable, segmentation maps can still provide accurate 2D hand contours.

{\noindent \bf Hand Depth Prior.} Depth maps provide information about the distance between the hands and the camera, helping to capture the relative positioning and spatial relationship of the hands in a real environment. Depth information is less affected by variations in lighting conditions, making hand understanding more reliable in environments with varying or poor lighting. As for the depth-scale ambiguity problem, our solution benefits from the human-centric vision foundation model Sapiens~\cite{khirodkar2024sapiens}, which is specifically optimized to effectively handle this challenge.

{\noindent \bf Fusion Alignment Encoder (FAE).}
A straightforward solution for integrating auxiliary 2D priors would be employing the original vision foundation model's encoder, but this would impose substantial computational overhead during inference. Instead, we propose an efficient alternative that maintains competitive performance while significantly reducing computational costs. To this end, we propose a lightweight fusion alignment encoder to learn the fused auxiliary information embeddings directly from the image by distilling the vision foundation models with MSE optimization, as shown in stage 1 of Fig. \ref{fig:pipline}. This approach allows us to bypass the need for additional foundation model encoders to obtain these prior features during inference. 

Given $\mathbf{F}_k$, $\mathbf{F}_m $, and $\mathbf{F}_d$ to represent the foundation model prior features of 2D key points, segmentation map and depth map, the fused prior feature $\mathbf{F}_p$ can be acquired by the learnable projection layer $Proj$. The formulation can be expressed as:
\begin{align}
\mathbf{F}_p =Proj(\mathbf{F}_k, \mathbf{F}_s, \mathbf{F}_d).
\end{align}

The output feature $\mathbf{F}_{fa}$ of the fusion alignment encoder will learn to align  $\mathbf{F}_p$.

{\noindent \bf Two-Hand Recovery Pipline.}
In our two-hand recovery framework, a transformer encoder effectively integrates image features $\mathbf{F}_i$ with fused prior features $\mathbf{F}_p$, followed by a hand regressor that predicts hand parameters from these unified representations. Then the final integrated feature $\mathbf{F}$ fed into the hand regressor can be expressed as:
\begin{align}
 \mathbf{F} = TransEnc(<\mathbf{F}_i, \mathbf{F}_p>){[0:l]},
\end{align}
Here, $<, >$ denotes the concat operation. $TransEnc$ represents the Transformer encoder. $l$ denotes the feature map’s channel length.

{\noindent \bf Two-Hand Recovery Loss Function.} Building on previous two-hand recovery approaches~\citep{moon2023bringinginterwild, yu2023acr}, We train our model in an end-to-end fashion by minimizing the L1 distance between the predicted and ground truth (GT) MANO parameters, the 3D and 2.5D joint coordinates, as well as the 3D relative translation. For training the fusion alignment encoder, we use MSE loss. Given feature $\mathbf{F}_{fa}$ from fusion alignment encoder, the total loss can be represented as:
\begin{align}
 \mathcal{L}_{total} = \mathcal{L}_{hand} + \mathcal{L}_{prior}{(\mathbf{F}_p, \mathbf{F}_{fa})}.
\end{align}

\subsection{Two-Hand Spatial Interactions Refinement}

In addition to 2D prior alignment, we argue that the reconstructed hands may still suffer from inconsistency in physical interactions where one hand occludes the important fingers of the other hand. In this scenario, the three types of additional information mentioned earlier are unable to provide effective guidance for the occluded parts of the hand. Consequently, the estimated occluded regions of both hands are prone to penetration issues.

To this end, we propose the two-hand diffusion model with penetration guidance refines hand interactions by iteratively denoising interpenetrated poses, gradually guiding them toward physically plausible configurations.

\begin{table*}
\centering
\resizebox{0.8\textwidth}{!}{\begin{tabular}{lccccccc}
\toprule
\multicolumn{1}{l}{\bf Methods} & \multicolumn{1}{c}{\bf MRRPE}& \multicolumn{1}{c}{\bf MPJPE}& \multicolumn{1}{c}{\bf MPVPE} & \multicolumn{1}{c}{\bf IH MPJPE}  & \multicolumn{1}{c}{\bf IH MPVPE} & \multicolumn{1}{c}{\bf SH MPJPE}    & \multicolumn{1}{l}{\bf SH MPVPE}  \\ \midrule

Moon et al. \citep{moon2020interhand2} & - & 13.98 & -& 16.02 & -& 12.16  & - \\
Zhang et al.  \citep{zhang2021interacting} &-  & 11.58  & 12.04  & 11.28 & 12.01 & 11.73 & 12.06 \\ 
IntagHand \citep{li2022interacting} & - & 9.95   & 10.29   & 10.27  & 10.53  & 9.67  & 9.91 \\
Zuo et al. \citep{zuo2023reconstructing} & - & 8.34 &8.51  & -  & - & - & -  \\
ACR \citep{yu2023acr} & - & 8.09 &8.29  & 9.08   & 9.31 & 6.85 & 7.01  \\
InterWild \citep{moon2023bringinginterwild} & 26.74 & 7.85   & 8.16   & 8.24 & 8.68  & 6.72 & 6.93 \\
Ren et.al \citep{ren2023decoupled} & 28.98 & 7.51   & 7.72  & -  & -  & -  & - \\
InterHandGen \citep{lee2024interhandgen} & 25.42 &  7.50  &  7.78  & 8.13 &  8.52 & 6.47 & 6.85 \\
4DHands \citep{lin20244dhands} & 24.58 & 7.49   & 7.72  & -  & -  & -  & - \\

\textbf{Ours} & \textbf{21.60}  & \textbf{5.36} &\textbf{5.58}   & \textbf{5.93}   & \textbf{5.87}  & \textbf{4.84}  & \textbf{4.86}

\\ \bottomrule
\end{tabular}}
\caption{Comparison with state-of-the-art methods on InterHand2.6M\citep{moon2020interhand2} 5fps test dataset. The results that are bolded and underlined represent the best result, while the bolded results represent the second-best result.}
\label{table:interhand}
\end{table*}

{\noindent \bf Two-Hand Penetration-Free Diffusion Model.} We implement a two-hand diffusion model to restore clear hand poses using corresponding penetrated reference two-hands as conditional input and incorporate penetration gradient guidance during the denoising phase. For penetrated two-hands, we generate them using two approaches: 1) the first involves synthesizing them with a low-performance two-hand estimation model and selecting the interpenetrated two-hand results. 2) the second applies slight noise to the ground truth MANO parameters of the hands until penetration occurs. 

Our method significantly differs from and holds advantages over InterHandGen~\cite{lee2024interhandgen} (using diffusion-based regularization for output) and zuo et al.~\cite{zuo2023reconstructing} (extracting interacting feature from CNN Encoder), as explicitly modeling interactive priors through a diffusion-based approach, effectively transform penetrated hands into their clean, collision-free counterparts. As shown in Stage 2 of Fig.~\ref{fig:pipline}, before using the two-hand diffusion model, we perform an IoU and penetration check between the two hands to reduce unnecessary diffusion inference in most cases. The gradient-guided two-hand diffusion effectively alleviates the penetration problem of the occluded regions.

{\noindent \bf Two-hand Diffusion Loss Function.} Our two-hand diffusion loss minimizes the L2 distance at each timestep between the clean hands $ \mathbf{X}_0 $ and the noisy hands $\mathbf{X}_t $ input to the model, conditioned on the timestep $t$ and the penetrated hand inputs $ \mathbf{X}_c$. Given two-hand diffusion model $\mathcal{D}$, the diffusion loss can be formulated as:
\begin{align}
 \mathcal{L}_{diffusion} = \| \mathbf{X}_0 - \mathcal{D}(\mathbf{X}_t, \mathbf{X}_c) \|_2 .
\end{align}

{\noindent \bf Collision Gradient Guidance.} During inference, we introduce a gradient-guided strategy to resolve hand-hand occlusion and prevent interpenetration. At each denoising step of the reverse diffusion process, we compute a collision loss between both hands and iteratively adjust hand poses via gradient descent. Specifically, clean two-hand parameters $\hat{\mathbf{X}}_0$ are first estimated from $\mathbf{X}_{t-1}$ using DDIM sampling. These parameters are fed into the MANO model to obtain mesh vertices $\mathbf{V}_{t-1}$ and $\mathbf{V}_c$. To accurately detect collisions, we design a hybrid distance-orientation criterion: 1) Calculate Chamfer distances $\mathbf{N}_{ij} = |\mathbf{V}_{t-1}^i - \mathbf{V}_c^j|^2$ and retain vertex pairs with $\mathbf{N}_{ij} < d_{\text{threshold}}$. 2) For retained pairs, compute the cosine similarity $\cos(\theta_{ij})$ between their normal vectors, identifying collisions when $\cos(\theta_{ij}) < \cos(\theta_{\text{thre}})$. This yields a collision set $\mathbf{C}_{\text{col}}$. We then formulate a robust collision loss using the GMoF function:
\begin{equation}
\begin{aligned}
  \mathcal{L}_{collision} &= \sum\limits_{i}\sum\limits_{j}(\frac{||\mathbf{V}_{t-1}^{i} - \mathbf{V}_{c}^j||^{2}}{||\mathbf{V}_{t-1}^{i} - \mathbf{V}_{c}^j||^{2} - \rho}),\\
\end{aligned}
\end{equation}
and update $\hat{\mathbf{X}}_0$ by propagating the negative gradient of this loss:
\begin{equation}
\begin{aligned}
  \hat{\mathbf{X}_{0}} &= \hat{\mathbf{X}_{0}} - \lambda(\delta_{i}\mathcal{L}_{collision}),
\end{aligned}
\end{equation}
where $\lambda$ controls the adjustment magnitude.

\begin{figure*}[th!]
\begin{center}
\includegraphics[width=.83\textwidth]{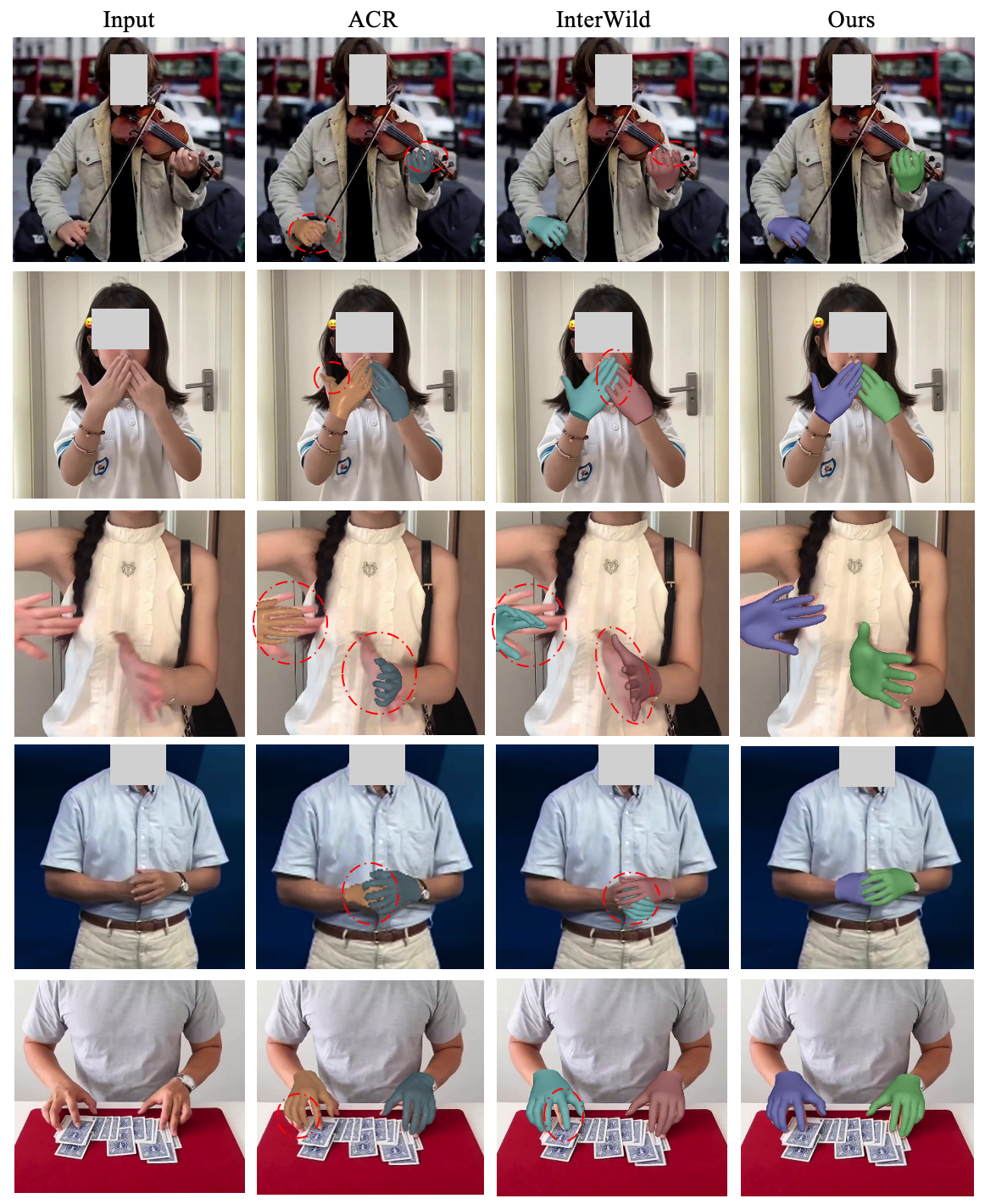}
\end{center}
\caption{ Qualitative two-hand recovery results in real scenes. 
The images are all sourced from the internet. The red circle indicates distortion or inaccurate estimation.}
\label{fig:vis_itw}
\end{figure*}

\section{Experiments}
This section validates our method's efficacy through quantitative and visual experiments, demonstrating: 1) the fusion alignment encoder's efficiency, and 2) the two-hand diffusion module's capability in eliminating interaction penetrations. Datasets and metrics details are provided in the Appendix.

\subsection{Implementation Details}
{\bf Two-Hand Reconstruction Model.} We implement our network using PyTorch~\citep{paszke2019pytorch}. For the image feature extractor, we use ResNet-50~\citep{he2016deep} as the backbone, while for 2D prior information encoders and fusion alignment encoder, we use the human-centric vision foundation model Sapiens~\cite{khirodkar2024sapiens} and ResNet-50. The hand bounding box detector utilizes RTMDet~\citep{lyu2022rtmdet}. Our model is trained on 4 A100 GPUs using the AdamW optimizer, starting with an initial learning rate of 1e-4, which is reduced by a factor of 10 at the 4th epoch. We use a mini-batch size of 48. For other details, we follow the approach in ~\citep{moon2023bringinginterwild}. Our training dataset only a few representative two-hand and single-hand datasets, including InterHand2.6M~\citep{moon2020interhand2}, Re:InterHand~\citep{moon2024dataset}, COCO whole-body~\citep{jin2020whole}, FreiHand~\citep{zimmermann2019freihand} and HO-3D~\citep{hampali2020honnotate}, which is less than the experimental setups of the latest methods 4DHands~\cite{lin20244dhands} (3 types of two-hand datasets and 9 types of one-hand datasets). For testing, we primarily use InterHand2.6M, FreiHAND and the in-the-wild dataset HIC~\citep{tzionas2016capturing}.

{\bf \noindent Two-Hand Diffusion Model.} We employ a transformer-based architecture for our two-hand diffusion model, utilizing MLPs to encode the input timesteps and fully connected layers to encode the interpenetrated two-hand inputs and predict the clean two-hand outputs. The diffusion model adopts an MDM-style~\cite{tevet2022human} diffusion process to enhance geometric learning. This model has been trained with 1,000 noising steps and a cosine noise schedule. The training datasets include InterHand2.6M~\citep{moon2020interhand2} and Re:InterHand ~\citep{moon2024dataset}.

\subsection{Datasets Details} The datasets are divided into two main categories: interacting hands (IH) and single hand (SH).
{\noindent \bf InterHand2.6M~\citep{moon2020interhand2}} features both precise human (H) and machine (M) 3D pose and mesh annotations, encompassing 1.36 million frames for training and 850,000 frames for testing.  
{\noindent \bf Re:InterHand~\citep{moon2024dataset}} consists of 739K video-based images and 493K frame-based images from third-person viewpoints, and 147K video-based images from egocentric viewpoints.
{\noindent \bf COCO WholeBody~\citep{jin2020whole}} extends the COCO dataset~\citep{lin2014microsoft} by adding comprehensive whole-body annotations. It includes manual annotations covering the entire human body.
{\noindent \bf FreiHand~\citep{zimmermann2019freihand}} is a dataset designed for single-hand 3D pose estimation, providing MANO annotations for each frame. It includes 4 × 32,560 frames for training and 3,960 frames for evaluation and testing.
{\noindent \bf HO-3D~\citep{hampali2020honnotate}} focuses on hand-object interactions, comprising 66,000 training images and 11,000 test images across 68 different sequences. 
{\noindent \bf HIC} provides diverse hand-hand interacting and object-hand interacting sequences and contains 3D GT meshes of both hands. It contains images with much more diverse and realistic appearances compared to InterHand2.6M.

\subsection{Evaluation Metrics} We mainly adopt Mean Per Joint Position Error (MPJPE) and Mean Per Vertex Position Error (MPVPE) to measure the 3D errors (in millimeters) of the pose and shape of each estimated hand after aligning them using a root joint translation, and Mean Relative-Root Position Error (MRRPE) to measure the performance of relative positions (in millimeters) of two hands.  Procrustes-aligned mean per joint position error (PA-MPJPE) and  Procrustes-aligned mean per vertex position error (PA-MPVPE) refer
to the MPJPE and MPVPE after aligning the predicted hand
results with the Ground Truth using Procrustes alignment,
respectively. To better investigate the impact of incorporating additional 2D information on performance, we introduce MPJPE-XY, MPJPE-Z, MPVPE-XY, and MPVPE-Z in the ablation study. These metrics calculate the hand recovery error of MPJPE and MPVPE relative to the ground truth in the XY and Z dimensions, respectively.

\begin{table}
    \centering
    \resizebox{0.8\linewidth}{!}{\begin{tabular}{lccccccc}
    \toprule
    \multicolumn{1}{l}{\bf Methods} & \multicolumn{1}{c}{\bf MRRPE}& \multicolumn{1}{c}{\bf MPJPE} & \multicolumn{1}{c}{\bf MPVPE}
    \\ \midrule
    
    IntagHand \citep{li2022interacting} & 73.04 & 20.38  & 21.56  \\
    InterWild \citep{moon2023bringinginterwild} & 26.43 & 15.62   & 15.17    \\
    4DHands \citep{lin20244dhands} & 25.26 &9.32   & 9.93  \\
    \textbf{Ours} & \textbf{22.24}  & \textbf{6.67} &\textbf{6.93}    \\

     \bottomrule
    \end{tabular}}
    \caption{Comparison with state-of-the-art methods on HIC dataset~\citep{tzionas2016capturing}.} 

    \label{table:hic}
\end{table}

\subsection{ Comparison with State-of-the-Art Methods}

{\noindent \bf Quantitative Results in InterHand2.6M Datasets.} 
We conduct a comprehensive comparison of our method with recent state-of-the-art (SOTA) hand pose and shape estimation methods on the InterHand2.6M test dataset, as presented in Table~\ref{table:interhand}. Our method achieves the best performance on MRRPE metric with 21.60mm, surpassing InterWild, Ren et al., and 4DHands by 5.14mm, 7.38mm, and 2.98mm respectively. Our method also demonstrates consistent improvement in MPJPE and MPVPE, outperforming the current best method, 4DHands, by 2.13mm and 2.14mm respectively.  Furthermore, we observe consistent performance gains in both the IH MPJPE/MPVPE and SH MPJPE/MPVPE metrics, highlighting the generalizability and robustness of our method for both single-hand and interacting hand estimation.

\begin{table*}[t]
\centering
\resizebox{0.8\textwidth}{!}{\begin{tabular}{lccccccc}
\toprule
\multicolumn{1}{c}{\bf Methods} & \multicolumn{1}{c}{\bf MRRPE} & \multicolumn{1}{c}{\bf MPJPE} & \multicolumn{1}{c}{\bf MPVPE} &
\multicolumn{1}{c}{\bf MPJPE-XY} &\multicolumn{1}{c}{\bf MPJPE-Z} & \multicolumn{1}{c}{\bf MPVPE-XY} & \multicolumn{1}{c}{\bf MPVPE-Z}
\\ \midrule

Baseline                         & 25.30  & 7.77 & 7.93 & 5.21  & 4.54 & 5.29 & 4.63 \\
\textbf{+ Key Points}          & 24.71  & 6.48 & 6.72 & 4.28  & 4.43 & 4.39 & 4.53  \\
\textbf{+ Segmentation Prior}      & 24.52  & 6.19 & 6.34 & 4.21  & 4.40 & 4.33 & 4.50  \\
\textbf{+ Depth Prior}             & 22.38  & 5.74 & 5.98 & 4.13  & 3.37 & 4.19 & 3.46   \\
\textbf{+ Penetration-Free Diffusion} & \textbf{21.60} & \textbf{5.36} & \textbf{5.58} &\textbf{3.87}  & \textbf{3.01} & \textbf{3.76} & \textbf{3.05}   \\ \bottomrule

\end{tabular}}
\caption{ Ablation studies on InterHand2.6M~\citep{moon2020interhand2}.} 

\label{table:ablation}
\end{table*}

{\noindent \bf Quantitative Results in HIC.} 
We present the results on the HIC dataset~\citep{tzionas2016capturing}, which features in-the-wild cross-hand data, to evaluate performance in real-world scenarios. The training sets for these models don't contain the HIC dataset.  In Table \ref{table:hic}, we compare these results with IntagHand, InterWild, and 4DHands, a state-of-the-art method specifically designed for two-hand recovery in the wild. Our method outperformed 4Dhands and InterWild across multiple metrics without using foundation model inference. These results highlight its superior stability on unseen data.

{\noindent \bf Qualitative Results in Real Scenes.} 
Fig.~\ref{fig:vis_itw} compares our method with ACR~\cite{yu2023acr} and InterWild~\cite{moon2023bringinginterwild} on real-world images. While existing methods exhibit misalignment (row 1), thumb distortion (row 2), penetration (rows 2,4), and failure under occlusion (row 3), our approach consistently delivers accurate and stable results across all challenging cases.

\begin{table}[t!]
\centering
\scriptsize
\begin{tabular}{lccccc}
\toprule
{\bf Methods}            &{\bf MRRPE} &{\bf MPJPE} &{\bf MPVPE} & {\bf Params} & {\bf FPS} \\
\midrule
$Ours^\cdot$    & 21.91  & 5.54 & 5.83 & 52.6M  + 1B      &    3      \\
$Ours^\star$    & 22.38 & 5.74 & 5.98    &   52.6M      &    56      \\
$Ours^{\star\star}$    & 21.60 & 5.36 & 5.58    &   74.2M      &    18\\      

\bottomrule
\end{tabular}
\caption{Comparison in model parameters and inference time. $\star$ represents with fusion alignment encoder \& without two-hand diffusion. $\star\star$ represents with fusion alignment encoder \&  two-hand diffusion. $\cdot$ denotes using foundation model encoder~\cite{khirodkar2024sapiens} without diffusion model for inference.} 
\label{table:fae}
\end{table}

\begin{figure}[t]
\begin{center}
\scriptsize
\includegraphics[width=.99\linewidth]{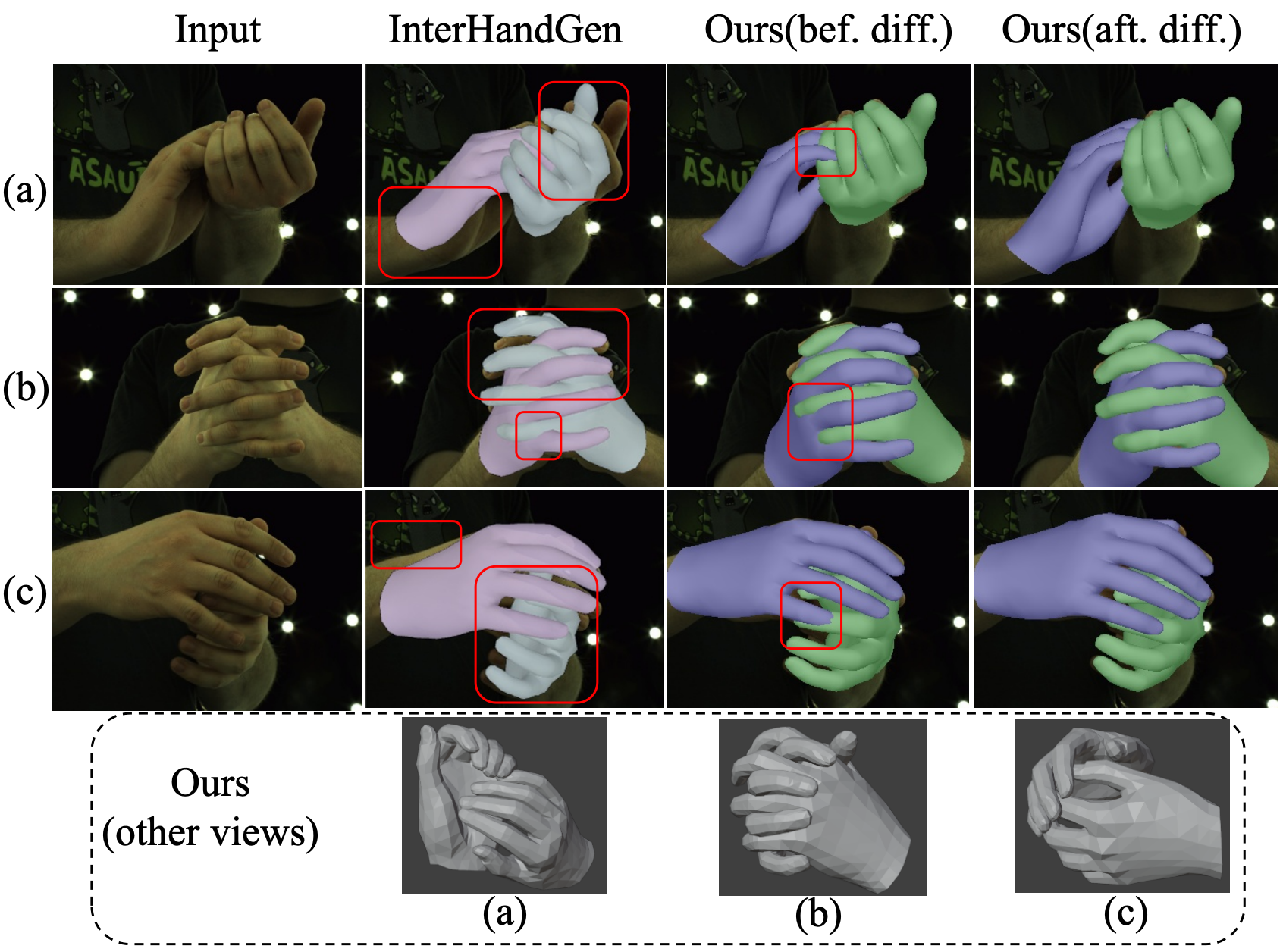}
\end{center}
\caption{Qualitative two-hand recovery results compared with InterHandGen~\cite{lee2024interhandgen}, Ours (before diffusion) and Ours (after diffusion) on InterHand2.6M~\cite{moon2020interhand2}.}
\label{fig:diff}
\end{figure}

\subsection{Ablation Study}
{\noindent \bf Efficiency of the Fusion Alignment Encoder.} 
In Table~\ref{table:fae}, we present a comprehensive comparison between our proposed fusion alignment encoder and the foundation model encoder~\cite{khirodkar2024sapiens} in terms of performance, model parameters and inference efficiency on a NVIDIA RTX 3090 GPU. Our method achieves an optimal balance across these three key metrics.

{\noindent \bf Effectiveness of Different Priors.} 
As shown in Table~\ref{table:ablation}, we gradually added different types of information for fusion to observe their impact on performance without using foundation model encoder in the inference stage. We found that incorporating 2D keypoints and segmentation maps significantly improved MPJPE and MPVPE, particularly in the XY dimension. Among them, 2D keypoints contributed more due to their detailed joint-level information, while segmentation maps provided coarser spatial cues. Additionally, fusing depth maps further improved MPJPE/MPVPE-Z and MRRPE, indicating that depth information enhances 3D structural reasoning.

 \begin{table}[t!]
 \centering
 \small
 \resizebox{0.8\linewidth}{!}{
\begin{tabular}{lccc}
\toprule
        Methods      & PenVol $\downarrow$      & PenDist $\downarrow$      & ProxRatio  $\uparrow$   \\
\toprule
InterHandGen \cite{lee2024interhandgen}  & 0.76          & 0.04          & 0.97          \\

\textbf{Ours} & \textbf{0.11} & \textbf{0.01} & \textbf{0.99}\\
\bottomrule
\end{tabular}}
\caption{Comparison on penetration metrics. We provide penetration metrics following \cite{lee2024interhandgen}: PenVol, PenDist, and ProxRatio stand for penetration volume, penetration distance, and proximity ratio, respectively.}
\label{table:pene}
\end{table}

{\noindent \bf Effectiveness of Two-Hand Diffusion Model.} 
Table~\ref{table:ablation} demonstrates the impact of the two-hand diffusion model on hand recovery performance. We can see that after adding diffusion, MRRPE, MPJPE, and MPVPE all achieve improvements and with the same improvement trend in both the XY and Z dimensions. As shown in Figure~\ref{fig:diff}, we present a visual comparison before/after applying the two-hand diffusion. Specifically, as an interaction prior, it effectively reduces occlusion-induced hand penetration. Table~\ref{table:pene} shows the improvement of our method in the de penetration metric compared to other approaches. These quantitative and visual results provide intuitive validation of the superiority of our method.

\section{Conclusion}
In this paper, we propose a two-hand reconstruction method that integrates additional 2D reference information to improve hand alignment and depth recovery performance. Furthermore, when one hand is occluded by another (making the 2D reference information for the occluded hand unreliable), we introduce a two-hand diffusion model as a interacting prior to address the penetration issue. Extensive qualitative and quantitative experimental results demonstrate that our method significantly outperforms previous two-hand and single-hand reconstruction approaches. \noindent{\bf Limitation and Future Work:} Our method still faces challenges in handling extreme motion blur in hand images, as the additional 2D information may become unreliable under such conditions. We believe that future integration of temporal processing could effectively alleviate this problem.
{
    \small
    \bibliographystyle{ieeenat_fullname}
    \bibliography{main}
}


\end{document}